**ARTICLE**

# Part-of-Speech Tagger for Bodo Language using Deep Learning approach

Dhrubajyoti Pathak, Sanjib Narzary, Sukumar Nandi, and Bidisha Som

Centre for Linguistic Science and Technology,
IIT Guwahati
drbj153@iitg.ac.in
Centre for Linguistic Science and Technology,
IIT Guwahati
sanjib_narzary@iitg.ac.in
Centre for Linguistic Science and Technology,
IIT Guwahati
sukumar@iitg.ac.in
Centre for Linguistic Science and Technology,
IIT Guwahati
bidisha@iitg.ac.in

**Abstract**
Language Processing systems such as Part-of-speech tagging, Named entity recognition, Machine translation, Speech recognition, and Language modeling (LM) are well-studied in high-resource languages. Nevertheless, research on these systems for several low-resource languages, including Bodo, Mizo, Nagamese, and others, is either yet to commence or is in its nascent stages. Language model plays a vital role in the downstream tasks of modern NLP. Extensive studies are carried out on LMs for high-resource languages. Nevertheless, languages such as Bodo, Rabha, and Mising continue to lack coverage. In this study, we first present BodoBERT, a language model for the Bodo language. To the best of our knowledge, this work is the first such effort to develop a language model for Bodo. Secondly, we present an ensemble DL-based POS tagging model for Bodo. The POS tagging model is based on combinations of BiLSTM with CRF and stacked embedding of BodoBERT with BytePairEmbeddings. We cover several language models in the experiment to see how well they work in POS tagging tasks. The best-performing model achieves an F1 score of 0.8041. A comparative experiment was also conducted on Assamese POS taggers, considering that the language is spoken in the same region as Bodo.

## 1. Introduction

Part-of-speech (POS) tagging is one of the building blocks of modern Natural Language Processing (NLP). POS tagging automatically assigns grammatical class types to words in a sentence, such as Noun, Pronoun, Verb, Adjective, Adverb, Conjunction, Punctuation, etc. Various approaches are there for automatic POS tagging; however, the Deep neural network approach has achieved state-of-the-art (SOTA) accuracy for resource-rich languages. POS tagging plays a vital role in various text processing tasks, such as Named entity recognition (NER) (Aguilar et al. 2019), Machine translation (MT) (Niehues and Cho 2017), Information extraction (IE) (Bhutani et al. 2019), Question Answering (QA) (Le-Hong and Bui 2018) and Constituency parsing (Shen et al. 2018). Therefore, a well-performing POS tagger is inevitable in developing a successful NLP system for a language.

Bodo (Boro) is a Tibeto-Burman[a] morphologically rich language, mainly spoken in Assam, a state in northeastern India. The Bodoland Territorial Region (BTR), an independent entity in Assam, uses it as its official language. The Devanagari script is used to write Bodo. As per the 2011 Indian census, Bodo has about 1.5 million speakers, making it the 20th most spoken language in India among the 22 scheduled languages. Even though most Bodo speakers are ethnic Bodos, Assamese, Rabha, Koch Rajbongshi, Santhali, Garo, and the Muslim community in the Bodoland Territorial Region also speak the language.

However, the state of the NLP systems for Bodo is in a very nascent stage. The study of fundamental NLP tasks in Bodo language, such as Lemmatization, Dependency parser, Language Identification, Language modeling, POS tagging, and NER has yet to start. While Word Embeddings or Language models play a crucial role in a Deep Learning approach, we observe that no existing pre-trained Language Models cover the Bodo language. As a result, we could not find any downstream NLP tools developed using a deep learning method.

With 1.5 million Bodo speakers, the need for developing NLP resources for the Bodo language is highly demanding. Motivated by this gap in research of the NLP areas of the Bodo language, we present the first language model for Bodo language- BodoBERT based on BERT architecture (Devlin et al. 2018). We prepare a monolingual training corpus to train the LM by collecting and acquiring corpus from various sources. After that, we develop a POS tagger for Bodo language by employing the BodoBERT.

We explore different sequence labeling architectures to get the best POS tagging model. We conduct three experiments to train the POS tagging models- a) Fine-tuning based, b) Conditional Random Field (CRF) based, and c) Long-short term memory (BiLSTM)-CRF based. Among the three, the BiLSTM-CRF-based POS tagging model performs better than the others.

Bodo language uses the Devanagari script, which the Hindi language also uses. Therefore, we conducted the POS tagging experiment with LMs that were trained in Hindi to compare the performance with the BodoBERT. We cover LMs such as Fasttext, Byte Pair Embeddings (BPE), Contextualise Character Embedding (FlairEmbedding), MuRIL, XLM-R, and IndicBERT. We employ two different methods to embed the words in the sentence to train POS tagging models using BiLSTM-CRF architecture- Individual and Stacked methods. In the Individual method, the model trained using BodoBERT achieves the highest F1 score of 0.7949. After that, we experiment with the Stacked method to combine the performance of the BodoBERT with other LMs. The highest F1 score in the stacked method reached 0.8041. We believe this is not only the first Neural Network-based POS tagging model for the language but also the first POS tagger in Bodo. Our contributions can be summarized as follows:

- Proposed a Language model for Bodo language based on the BERT framework. To the best of our knowledge, this is the first Language model for Bodo.
- Presented comparison of different POS tagging models for Bodo by employing state-of-the-art sequence tagger frameworks such as CRF, Fine-tuned LMs, and BiLSTM-CRF.

---

[a] The Sino-Tibetan language family includes the Tibeto-Burman languages. Tibeto-Burman language does not include Chinese languages.

- Comparison of POS tagging performance of different LMs in POS tagging models such as Fasttext, BPE, XLM-R, FlairEmbedding, IndicBERT, and MuRIL embedding using Individual and Stacked embedding methods.
- The top-performing Bodo POS tagging model and BodoBERT are made publicly available [b].

This paper is organized as follows- Section 2 describes related works on POS tagging for similar language. The details about the pre-training of BodoBERT and Bodo corpus used in training are presented in Section 3. Section 4 presents the experiments carried out to develop the Neural Network POS tagger. The Section also includes the description of the annotated dataset and POS tagset in the experiment. In Section 5, we present all the experiment results of all three models using different sequence tagging architectures. Finally, we conclude our paper in Section 6.

## 2. Related Work

This section presents related works about POS tagging in different languages. In our literature study, we do not find any prior work on Language Modeling for Bodo Langauge. Furthermore, to the best of our knowledge, there is no previous work on POS tagging for Bodo Language. Therefore, we discuss recent research works reported on neural network-based POS tagging in various low-resource languages belonging to the northeastern part of India.

In paper (Pathak et al. 2022), an Assamese POS tagger based on Deep Learning (DL) approach is proposed. Various DL architectures are used to develop the POS model, which includes CRF, Bidirectional Long Short-term Memory (BiLSTM) with CRF, and Gated Recurrent Unit (GRU) with CRF. The BiLSTM-CRF model achieved the highest tagging F1 score of 86.52. Pathak et al. (2023) presented an ensemble POS tagger for Assamese, where two DL-based taggers and a rule-based tagger are combined. The ensemble model achieved an F1 score of 0.925 in POS tagging. Warjri et al. (2021) presented a POS-tagged corpus for Khasi language. The paper also presented a DL-based POS tagger for the language using different architectures, including BiLSTM, CRF, and Character-based embedding with BiLSTM. The top performing tagger achieved an accuracy of 96.98% using BiLSTM with CRF technique. Alam et al. (2016) proposed a neural network-based POS tagging system for Bangla using BiLSTM and CRF architecture. They also used a pre-trained word embedding model during training. The model achieved an accuracy of 86.0%. In paper (Pandey et al. 2022), reported work on POS tagging of the Mizo language. They employed classical LSTM and Quantum-enhanced Long short-term memory (QLSTM) in training and reported a tagging accuracy of 81.86% using LSTM network. Kabir et al. (2016) presented their research on POS tagger for Bengali language using a DL-based approach. The tagger achieves an accuracy of 93.33% using Deep Belief Network (DBN) architecture. There are other works that have been reported over the years on POS tagging for a variety of languages, including Assamese, Mizo, Manipuri, Khasi, and Bengali. However, they are based on traditional methods such as Rule-based and HMM-based models.

---

[b] https://anonymous.4open.science/status/BodoPoS-4C10

## 3. BodoBERT

The Transformer-based language model BERT achieves state-of-the-art performance on many NLP tasks. However, the success of BERT models on downstream NLP tasks is mostly limited to high-resource languages such as English, Chinese, Spanish, Arabic, etc. The BERT model is trained in English (Devlin et al. 2018). After that, the multilingual pre-trained models were released for 104 languages. However, Bodo language is not covered. We could not find any pre-trained language models that cover the Bodo language. So, this motivates us to develop a language model for Bodo using the vanilla BERT (Devlin et al. 2018) architecture.

### *3.1 Bodo raw corpus*

A large monolingual corpus is required to train a well-performed language model like BERT. Moreover, training BERT is a computationally intensive task, and it requires substantial hardware resources. On the other hand, getting decent monolingual raw corpus for Bodo has been an enduring problem for NLP research community. Although Bodo has a rich oral literary tradition, however, there was no standard writing script for writing until the year 2003. After a long history of the Bodo script movement, the Bodo language is recognized as a scheduled language of India by the Government of India, and the Devanagari script for writing is officially adopted.

We have curated the corpus after acquiring it from the Linguistic Data Consortium for Indian Languages (LDC-IL) (Ramamoorthy et al. 2019; Choudhary 2021). The text of the raw corpus is from different domains such as Aesthetics (Culture, Cinema, Literature, Biographies, and Folklore), Commerce, Mass media (Classified, Discussion, Editorial, Sports, General news, Health, Weather, and Social), Science and Technology (Agriculture, Environmental Science, Textbook, Astrology, Mechanical Engineering, and Environmental Science) and Social Sciences (Economics, Education, Political Science, Linguistics, Health and Family Welfare, History, Text Book, Law, etc). We also acquired another corpus from the work (Narzary et al. 2022). The final consolidated corpus has 1.6 million tokens and 191k sentences.

### *3.2 BodoBERT model training*

The architecture of the BodoBERT model is based on a multi-layer bidirectional Transformer framework (Vaswani et al. 2017). We use the BERT framework (Devlin et al. 2018) to train the LM using the masked language model and next-sentence prediction tasks. WordPiece tokenizer (Wu et al. 2016) is used for embeddings with 50,000 vocabularies. The BERT model architecture is described in the guide "The Annotated Transformer"[c] and the implementation details are provided in "Tensorflow code and BERT model"[d].

The model is trained with six layers of transformers block with a hidden layer size of 768 and the number of self-attention heads as 6. There are approximately 103M parameters. The model was trained on Nvidia Tesla P100 GPU (3584 Cuda Cores) for 300K steps with a maximum sequence length of 128 and batch size of 64. We used the Adam optimizer (Kingma and Ba 2014) with a warm-up of the first 3000 steps to a peak learning rate of 2e-5. The pre-training took approximately seven days to complete.

---

[c] http://nlp.seas.harvard.edu/2018/04/03/attention.html
[d] https://github.com/google-research/bert

## 4. Bodo POS Tagger

Part-of-speech tagging belongs to the sequence labeling tasks. There are different stages in developing a DL-based POS tagger. This section presents various stages of development of the Bodo POS tagger.

### *4.1 Annotated Dataset*

A DL-based method requires a large size of properly annotated corpus to train a POS tagging model. The performance of a well-performed tagging model depends upon the quality of the annotated dataset. On the other hand, the availability of annotated corpus in the public domain is very rare. Moreover, it is also a tedious and time-consuming task to annotate a corpus manually. Therefore, it is a challenging task to build a deep learning model for a low-resource language. In our literature study, we could find only one Bodo annotated corpus - Bodo Monolingual Text Corpus (ILCI-II 2020b). The corpus is tagged by language experts manually as part of the project Indian Languages Corpora Initiative Phase–II (ILCI Phase- II), initiated by the Ministry of Electronics and Information Technology (MeitY), Government of India, and Jawaharlal Nehru University, New Delhi. The corpus contains approximately 30k sentences and 240k tokens comprised of different domains. The statistics of the annotated dataset are reported in Table 1. The corpus is tagged according to the Bureau of Indian Standards (BIS) tagset. We use this dataset for all our experiments to train neural-based taggers for Bodo.

**Table 1.** Statistics of Bodo POS annotated dataset

| File Name | Sentence Count | Token Count |
| --- | --- | --- |
| Training set | 24003 | 192k |
| Dev set | 2325 | 23k |
| Test set | 3161 | 23k |

We prepare the tagged dataset in CoNLL-2003 (Sang and De Meulder 2003) column format in which each line contains one word, and the corresponding POS tag is separated by tab space. An empty line represents the sentence boundary in the dataset. In Table 2, the sample column format is shown.

For training, we first randomized the dataset; after that, the dataset was divided into 80:10:10 train, development, and test sets, respectively.

### *4.2 POS taggset*

The TDIL dataset is tagged according to the Bureau of Indian Standards (BIS) tagset, which is considered the national standard for annotating Indian languages. The dataset contains eleven (11) top-level categories that include 34 tags. The complete tagset used in our experiment is reported in Table 3.

### *4.3 Language Model*

Language models are a crucial component in a deep learning-based model. These models are typically trained on large unlabeled text corpus to capture both semantic and syntactic meanings. The size of the training corpora (Bojanowski et al. 2017) impacts the quality of the language models. In our literature survey, we could not find any other LMs that covered the Bodo language. The lack of availability of text corpus is one of the main factors behind this. Consequently, there exist no other LMs that are trained on Bodo except our newly trained BodoBERT. Therefore, to accomplish our comparative experiment, we consider pre-trained language models for Hindi as it shares the same written script with Bodo-Devanagari. Table 4 provides the details about the LMs that are used for training the Bodo POS tagging model. A brief description of these models is described below.

**Table 2.** Dataset format

| Word | Tag |
|------|-----|
| बियो | PR_PRP |
| 88 | QT_QTC |
| सानावनो | N_NST |
| सानखौ | N_NN |
| खेबसे | QT_QTC |
| गिदिंखनो | V_VM |
| ( | RD_PUNC |
| Revolution | RD_RDF |
| ) | RD_PUNC |
| । | RD_PUNC |

**FastText embedding** (Bojanowski et al. 2017) uses sub-word embedding technique and skip-gram method. It is trained on character *n*-grams of words to get the internal structure of a word. Therefore, it has the ability to get the word vectors for out-of-vocabulary (OOV) words by using the sub-word information from the previously trained model.

**Byte-Pair embedding** (Heinzerling and Strube 2018) are pre-computed on sub-word level. They can embed by splitting words into subwords or character sequences, looking up the pre-computed subword embeddings. It has the capability to deal with unknown words and has the ability to infer meaning from unknown words.

**Flair Embedding** (Akbik et al. 2018) is a type of character level embedding. It is a contextualized word embedding that captures word semantics in context, meaning that a word has different vector representations under different contexts. The embedding method based on recent advances in neural language modeling (Sutskever et al. 2014; Karpathy et al. 2015) that provides sentences to be modeled as distributions over sequences of characters instead of words (Sutskever et al. 2011; Kim et al. 2016).

**Multilingual Representations for Indian Languages (MuRIL)** (Khanuja et al. 2021) is a multilingual language model based on BERT architecture. It is pre-trained in 17 Indian languages.

---

[e] https://fasttext.cc/docs/en/pretrained-vectors.html
[f] https://github.com/bheinzerling/bpemb
[g] https://anonymous.4open.science/status/BodoPoS-4C10
[h] https://github.com/flairNLP/flair/blob/master/resources/docs/embeddings/FLAIR_EMBEDDINGS.md
[i] https://huggingface.co/google/muril-base-cased
[j] https://tinyurl.com/XLM-R-Embed
[k] https://indicnlp.ai4bharat.org/indic-bert/

7**Table 3.** Tagset used in the dataset

| S.No | Category | Type | Tag |
|---|---|---|---|
| 1 | Noun | Proper Noun | N_NNP |
| | | Noun (Location) | N_NST |
| | | Noun (unclassified) | N_NN |
| 2 | Pronoun | Personal | PR_PRP |
| | | Reflexive | PR_PRF |
| | | Reciprocal | PR_PRC |
| | | Relative | PR_PRL |
| | | Wh-words | PR_PRQ |
| | | Indefinite | PR_PRI |
| 3 | Demonstrative | Deictic | DM_DMD |
| | | Relative | DM_DMR |
| | | Wh-words | DM_DMQ |
| | | Indefinite | DM_DMI |
| 4 | Verb | Auxiliary Verb | V_VAUX |
| | | Main Verb | V_VM |
| | | Finite | V_VM_VF |
| | | Non-Finite | V_VAUX_VNF |
| 5 | Adjective | Adjective | JJ |
| 6 | Adverb | | RB |
| 7 | Post Position | | PSP |
| 8 | Conjunction | Conjunction | CC_CCD |
| | | Co-ordinator | CC_CCS |
| 9 | Particles | Classifier | RP_RPD |
| | | Interjection | RP_INJ |
| | | Negation | RP_NEG |
| | | Intensifier | RP_INTF |
| 10 | Quantifiers | General | QT_QTF |
| | | Cardinals | QT_QTC |
| | | Ordinals | QT_QTO |
| 11 | Residuals | Foreign word | RD_RDF |
| | | Symbol | RD_SYM |
| | | Punctuation | RD_PUNC |
| | | Echowords | RD_ECH |
| | | Unknown | RD_UNK |

**XLM-R** (Conneau et al. 2020) uses self-supervised training techniques in cross-lingual understanding, a task in which a model is trained in one language and then used with other languages without additional training data.

**IndicBERT** (Kakwani et al. 2020) based on Fasttext-based word embedding and ALBERT-based language models for 11 languages trained on the IndicCorp dataset.

**Table 4.** Training datasize of different language models

| Language model | Trained Corpus | Trained Datasize |
|---|:---:|:---:|
| FastTextEmbeddings (Bojanowski et al. 2017)[e] | Wiki | < 29M |
| Byte Pair (Heinzerling and Strube 2018)[f] | Wiki | 29M |
| BodoBERT (Bodo) [g] | Bodo corpus (Narzary et al. 2022) | 1.6M |
| Flair Embeddings (Akbik et al. 2018) [h] | Wiki+OPUS | ≈ 29M |
| MuRIL (Khanuja et al. 2021) [i] | CommonCrawl + Wiki | 788M |
| XLM-R Embedding (Conneau et al. 2020) [j] | CC-100 corpus | 1.7B |
| IndicBERT (Kakwani et al. 2020) [k] | Scraping | 1.84B |

### *4.4 Experiment on POS models*

In this section, we describe the experiments conducted to develop POS tagging models using different LMs, including BodoBERT. The experiment can be divided into three phases. In the first phase, we employ three different sequence labeling architectures that are shown to be a well performer in other languages, namely - Fine-tuning BodoBERT model for POS tagging, CRF (Lafferty et al. 2001) and the third one with BiLSTM-CRF (Hochreiter and Schmidhuber 1997; Rei et al. 2016). The pre-trained BodoBERT is used to get the embedding vector of the words present in the sentences during training with CRF and BiLSTM-CRF (Huang et al. 2015) architecture. It is observed that the BiLSTM-CRF-based model outperforms the other two models.

**Table 5.** Performance of POS tagging model in different methods

| Embeddings | Tagging model | F1-score(Micro) | F1 score (Weighted) |
|---|---|---|---|
| BodoBERT | CRF | 0.7583 | 0.7454 |
|  | Fine-tuned BERT | 0.7754 | 0.7775 |
|  | **BiLSTM + CRF** | **0.7949** | **0.7898** |

Therefore, the BiLSTM-CRF architecture is used in the second phase to develop POS tagging models employing different language models, including BodoBERT.

**Table 6.** The F1 score of different Language models in POS tagging task on Bodo and Assamese language

| Embeddings | Bodo | Assamese |
|---|---|---|
| FastTextEmbeddings | 0.7686 | 0.6981 |
| BytePairEmbeddings | 0.7669 | 0.7099 |
| **BodoBERT** | **0.7949** | 0.7033 |
| FlairEmbeddings (Multi) | 0.7885 | 0.7076 |
| MuRIL | 0.7708 | 0.7286 |
| XLM-R | 0.7638 | 0.7001 |
| IndicBert | 0.7235 | **0.7293** |

Apart from Bodo, we also conducted the same experiment on Assamese, another low-resource language. Existing Assamese pre-trained LMs are used for the experiment. To conduct the training for Assamese POS model, we acquired dataset (ILCI-II 2020a) from TDIL, which was tagged by language experts manually. There are 35k sentences from different domains and 404k words in the POS datasets. The Assamese annotated dataset also follows the BIS tagset and contains 41 tags with 11 top-level categories.

In the third phase, we further experiment with the Stacked embedding method to develop the Bodo POS tagging mode. The Stacked method allows us to learn how well one embedding performs when combined with others during the training process. The top-performing LM (BodoBERT) in the second phase is selected for further training. In the Stacked method, each one of the LMs is combined with BodoBERT to get the final word embedding vector.

Compared to the best Individual method, the Stacked embedding approach using BiLSTM-CRF improves the performance score for POS tagging by around 2%-7%. The model with **BodoBERT+BytePairEmbedding** attains the highest F1 score of 0.8041. The results of the experiment are listed in Table 7. The POS model architecture is illustrated in Figure 1. In all experiments, the Flair framework[1] is used to train the sequence tagging model.

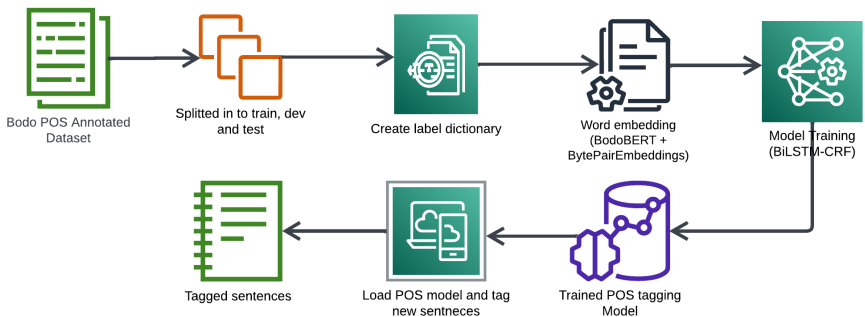

**Figure 1.** Block diagram of POS tagging model

**Table 7.** POS tagging performance in the stacked method using BiLSTM-CRF architecture

| Stacked Embeddings | F1 score |
| --- | --- |
| BodoBERT + FastTextEmbeddings | 0.7928 |
| **BodoBERT + BytePairEmbeddings** | **0.8041** |
| BodoBERT + mBERT | 0.799 |
| BodoBERT + FlairEmbeddings | 0.801 |
| BodoBERT + MuRIL | 0.785 |
| BodoBERT + XLM-R | 0.8003 |
| BodoBERT + IndicBert | 0.793 |

We explored different hyperparameters to optimize the configurations concerning the hardware constraint. After that, we use the same set of hyperparameters in all three experiments. We use a fixed mini-batch size of 32 to account for memory constraints. The early stopping technique is used if there is no improvement in the validation data accuracy. We use the Learning Rate Annealing factor for early stopping.

---

[1] https://github.com/flairNLP/flair

## 5. Result and Analysis

In this section, we present an analysis of our experiment and the performance of taggers. We evaluated the performance of the three sequence tagging methods: CRF, Fine-tuning of BodoBERT, and BiLSTM-CRF by measuring the micro F1 score. The weighted average of the tagging performance is also reported.

Table 5 illustrates an overview of the tagging performance in F1 score of the three models. We observe that the BiLSTM-CRF based model with BodoBERT performs the best with an F1 score of 0.7949 and Weighed the average F1 score as 0.7898. In contrast, the Fine-tune-based and CRF-based tagging model achieves an F1 score of 0.7754 and 0.7583, respectively. The performance comparison result of different LMs is reported in Table 6. We observe BodoBERT outperforms all the other pre-trained models. The Flair Embedding model achieves almost similar performance in tagging with an F1 score of 0.7885.

We also cover a similar low-resource language, Assamese spoken in the same region. The same set of experimental setups is used for the experiment. It provides us with an overview of how the models work on similar languages with almost the same size as the annotated dataset. It has been observed that the highest F1 score achieved in the case of Assamese is 0.7293 using IndicBERT. This is almost $\approx 7$ less than the highest of the Bodo POS tagging model. It could be because of the difference in the number of tagsets used in Assamese (41 tags) versus Bodo (34 tags).

**Data Augmentation experiment**: The overall best-performing model (BodoBERT+BytePairEmbedding) is employed to annotate a new set of sentences from another corpus. The corpus is taken from the Bodo corpus created by (Narzary et al. 2022). The annotated dataset is further manually corrected. The new dataset that comprises 10k is added to the existing training dataset. In order to evaluate the performance of the model when increasing the training dataset, the same architectures (BodoBERT+BytePairEmbedding in BiLSTM-CRF) are employed in further training the POS model with the same set of parameters. The test and dev are kept the same. It is observed that the model performance is enhanced by 1$\approx$2%. The model achieves an F1 score of 0.8494.

The tag-wise performance score of precision, recall, and F1 score, along with support consolidated micro, macro, and weighted score, are reported in Table 8. The learning curve of training and validation for the best-performing model is shown in Figure 2.

The learning curve implies that the dataset needs improvement as the training dataset does not provide sufficient information to learn the sequence tagging problem relative to the validation dataset used to evaluate it. Whatsoever, we get the first POS tagging model for the Bodo language. Eventually, it becomes the de facto baseline for the Bodo tagging.

The reference sentences and their corresponding tagged results from the proposed tagger are given below.

- <u>Reference sentences</u>

    Sentence 1: तिकेन $<N\_NNP>$ बर'आ $<N\_NNP>$ सासे $<QT\_QTC>$ मोजां $<JJ>$ फोरोंगिरि $<N\_ANN>$ । $<RD\_PUNC>$
    IPA: tikɛn boroɑ sase mojaŋ foroŋgiri
    English: *Tiken Bodo is a good teacher*
    Sentence 2: बडलेन्ड $<N\_NNP>$ मुलुगसोलोंसालिआ $<N\_NNP>$ क्रक्राझारआव $<N\_NNP>$ दं $<V\_VAUX>$ । $<RD\_PUNC>$
    IPA: boroland mulugɔsoloŋsaliɑ kokrɑjhɑrɑɔ

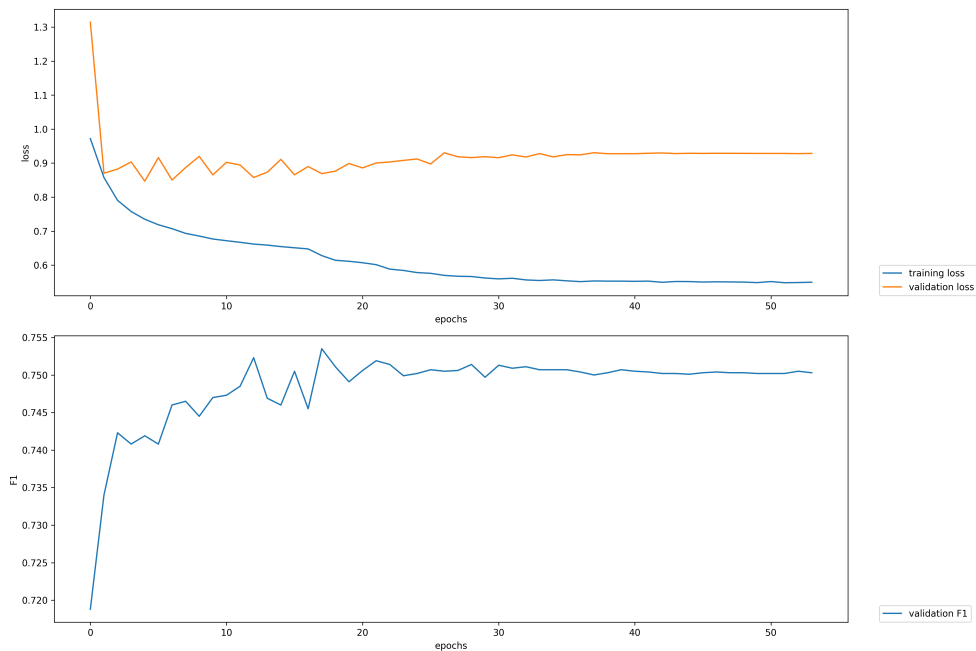

**Figure 2.** Learning curve of BodoBERT + BytePairEmbeddings based POS model

                dɔŋ
English:    *Bodoland University is situtated in Kokrajhar*

- Tagged by the proposed DL-based tagger

    Sentence 1: तिकेन $<N\_NNP>$ बर'आ $<N\_NN>$ सासे $<QT\_QTC>$ मोजां $<JJ>$ फोरोंगिरि $<N\_NN>$ । $<RD\_PUNC>$
    IPA:      tikɛn          boroa          sase          mojaŋ foroŋgiri
    English:    *Tiken Bodo is a good teacher*
    Sentence 2: बडलेन्ड $<N\_NNP>$ मुलुगसोलोंसालिआ $<N\_NN>$ कक्राझारआव $<N\_NN>$ दं $<V\_VAUX>$ । $<RD\_PUNC>$
    IPA:      boroland      mulugɔsoloŋsaliɑ      kokrɑjhɑrɑɔ dɔŋ
    English:    *Bodoland University is situtated in Kokrajhar*

In the above example, the word 'बर'आ' /boroɑ/ ('Bodo') in sentence 1, and मुलुगसोलोंसालिआ /mulugɔsoloŋsalia/ (University + nominative marker '-a') and कक्राझारआव /kokrɑjhɑrɑɔ/ (Kokrajhar + locative marker '-aɔ') in sentence 2 are proper nouns (*N_NNP*). But the tagger considers them as NOUN (generalized form) and tags them accordingly as *N_NN*. Likewise, फोरोंगिरि /foroŋgiri/ (Teacher) is an abstract noun (N_ANN). However, it is tagged as N_NN by the tagger. In other cases, the POS tagged the words correctly. If we consider only the top-level tag, the tagger performance increases.

**Table 8.** Tag-wise performance of best performing Bodo POS tagging model

| Tags | Precision | Recall | F1 score | Support |
|---|---|---|---|---|
| N_NN | 0.7439 | 0.7826 | 0.7628 | 11560 |
| V_VM | 0.8945 | 0.9366 | 0.9150 | 9005 |
| RD_PUNC | 0.9927 | 0.9960 | 0.9944 | 6815 |
| N_NNP | 0.7180 | 0.6727 | 0.6946 | 5264 |
| JJ | 0.6665 | 0.5554 | 0.6059 | 1975 |
| N_NST | 0.5703 | 0.5194 | 0.5436 | 1421 |
| CC_CCD | 0.9059 | 0.9634 | 0.9338 | 1039 |
| DM_DMD | 0.9699 | 0.9593 | 0.9646 | 908 |
| PSP | 0.6492 | 0.7624 | 0.7012 | 665 |
| QT_QTC | 0.7319 | 0.8743 | 0.7968 | 565 |
| RB | 0.7295 | 0.4958 | 0.5904 | 593 |
| PR_PRP | 0.8584 | 0.8491 | 0.8537 | 464 |
| RD_UNK | 0.6613 | 0.0861 | 0.1524 | 476 |
| RD_ECH | 0.2155 | 0.4266 | 0.2864 | 143 |
| QT_QTF | 0.3673 | 0.1949 | 0.2547 | 277 |
| PR_PRI | 0.6552 | 0.6683 | 0.6617 | 199 |
| CC_CCS | 0.4789 | 0.5574 | 0.5152 | 122 |
| RP_INTF | 0.3699 | 0.4154 | 0.3913 | 65 |
| V_VAUX | 0.4493 | 0.5082 | 0.4769 | 61 |
| PR_PRF | 0.1205 | 0.2222 | 0.1562 | 45 |
| RD_SYM | 0.9600 | 0.6486 | 0.7742 | 74 |
| QT_QTO | 0.5758 | 0.7308 | 0.6441 | 52 |
| DM_DMI | 0.2292 | 0.1864 | 0.2056 | 59 |
| RD_RDF | 1.0000 | 0.9455 | 0.9720 | 55 |
| PR_PRC | 0.1379 | 0.2963 | 0.1882 | 27 |
| PR_PRL | 0.5000 | 0.1346 | 0.2121 | 52 |
| PR_PRQ | 0.3704 | 0.3571 | 0.3636 | 28 |
| RP_RPD | 0.0400 | 0.0526 | 0.0455 | 19 |
| DM_DMQ | 0.4615 | 0.8000 | 0.5854 | 15 |
| RP_NEG | 0.0000 | 0.0000 | 0.0000 | 5 |
| RP_INJ | 0.0000 | 0.0000 | 0.0000 | 4 |
| DM_DMR | 0.0000 | 0.0000 | 0.0000 | 4 |
| micro avg | 0.8041 | 0.8041 | 0.8041 | 42056 |
| macro avg | 0.5320 | 0.5187 | 0.5076 | 42056 |
| weighted avg | 0.8021 | 0.8041 | 0.7990 | 42056 |

The reported highest score is arguably lower than the state-of-the-art score on resource-rich languages. This could be due to a variety of factors.

(1) The size of the annotated corpus may not be adequate.

(2) The training data size of BodoBERT may not be sufficient enough to capture the linguistic features of the language.
(3) The language model BodoBERT may need more improvement in capturing the linguistic characteristics of the Bodo language.
(4) The assignment of the tags to the words in the dataset may need some correction.

**Observation from confusion matrix**: The confusion matrix of the top-performing Bodo POS tagging model is reported in Figure 3. The confusion matrix covers the top twelve POS tags in terms of frequency counts in the test set. The diagonal entries of the matrix represent the correctly predicted tags, and the off-diagonal entries represent incorrectly classified tags.

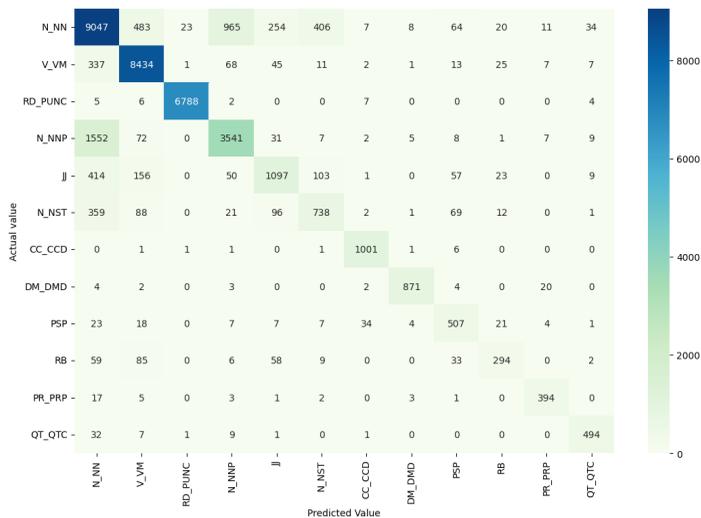

**Figure 3.** Confusion matrix of BodoBERT + BytePairEmbeddings POS model

It is observed that the common error occurs in Noun (N_NN), Proper Noun (N_NNP), Locative Noun (N_NST), VERB (V_VM), Adjective (JJ), and Adverb (RB) in the taggers. We can draw the following observation from the confusion matrix.

- The most common error in the dataset is intra-class confusion, i.e., confusing Noun (N_NN) with Proper (N_NNP) and Noun (N_NN) with Locative noun (N_NST). This might have occurred due to the similarity in the attached features of nouns, pronouns, and locative nouns.
- In some instances, the tagger incorrectly predicts a noun when its class type changes to an adjective in a sentence. It happens when it describes another noun, e.g., *'Bodo language'*; in this case, although *Bodo* is a proper noun, it is being used to describe the noun- *language*. Therefore it becomes an adjective.
- Sometimes, it becomes difficult to figure out the correct tag- JJ or N_NN, V_VM or N_NN, and RB or N_NN. So, many times the tagger incorrectly tags as N_NN for JJ, V_VM, and RB.
- Furthermore, Bodo has no similar orthographic conventions to differentiate the proper nouns as done using capitalization in English. Therefore, it is difficult for a machine to differentiate a proper noun from other nouns.

## 6. Conclusion

In this work, we presented a language model, BodoBERT, for the Bodo language based on BERT architecture. We develop a POS tagging model employing three distinct sequence tagging architectures using the BodoBERT. Upon applying the BodoBERT language model in POS tagging, we obtained two outcomes: first, an evaluation of the pre-trained BodoBERT's performance on a downstream task in different architectures; Second, we obtained a model for POS tagging in the Bodo language. We also compared the performance of BodoBERT to that of other LMs using two methods: Individual and Stacked. The Stacked method improves the performance of the POS tagging model. In our experiment, the model that uses BodoBERT and BytePairEmbeddings together in a stacked method does better.

Despite the fact that the Bodo POS tagger is unable to attain state-of-the-art accuracy in comparison to resource-rich languages, we feel that our POS model can serve as a baseline for future studies. Our contributions may be useful to the research community in terms of using the language model BodoBERT, and the POS tagging model for various downstream tasks.